\documentclass[12pt,journal,compsoc,onecolumn]{IEEEtran}

\usepackage{cite}
\usepackage{amsmath,graphicx}
\def\BibTeX{{\rm B\kern-.05em{\sc i\kern-.025em b}\kern-.08em
    T\kern-.1667em\lower.7ex\hbox{E}\kern-.125emX}}

\usepackage{amsmath,amssymb,amsfonts}
\usepackage{algorithmic}
\usepackage{graphicx}
\usepackage{textcomp}
\usepackage{xcolor}

\usepackage{pifont}
\usepackage{multirow}

\newtheorem{proposition}{\bf Proposition}

\let\OLDthebibliography\thebibliography
\renewcommand\thebibliography[1]{
  \OLDthebibliography{#1}
  \setlength{\parskip}{3.0pt}
  \setlength{\itemsep}{0pt plus 0.1ex}
}

\title{Cayley-Reparameterized Graph Convolutional Networks for Active Learning}
\author{Hichem Sahbi \\
$ $ \\
Sorbonne University, CNRS, LIP6,  F-75005, Paris, France}

\begin{document}

\maketitle

\begin{abstract}
The efficacy of skeleton-based action recognition is  contingent upon the availability of massive, expert-annotated datasets --- a requirement that remains a bottleneck for real-world deployment. While Active Learning (AL) offers a mechanism for label efficiency, traditional pool-based methods are limited by the discrete, often rigid, nature of candidate data selection. This paper introduces an exemplar design framework that shifts the AL paradigm from pool-based selection to the full design of optimal exemplars  for annotation. We propose a model that identifies the most informative exemplars as the fixed-point solution to a unified objective function balancing classifier uncertainty and exemplar diversity. To ensure these designed exemplars are valid and in-distribution, we consider a Cayley-reparameterized graph convolutional network (GCN) that yields a  mapping between input skeletal data  and a  latent space. This allows for a more tractable optimization of exemplars in the latent domain and their stable,  bijective back-projection to the ambient space.  Experiments conducted on standard benchmarks show that our framework significantly outpaces related  AL selection strategies, achieving higher accuracy with a very small fraction of the labeling budget. 
\end{abstract}

\section{Introduction}
\label{sec:intro}
Skeleton-based  action recognition has emerged as a cornerstone of visual information processing, enabling the  analysis of human body and hand interactions through joint-coordinate modeling~\cite{Du2015,Zhua2016,RiemannianManifoldTraject,DeepGRU,Jiang2020,Liu2021}. By abstracting motion into a topological graph, this modality remains uniquely robust in cluttered or poorly illuminated environments where traditional RGB-based methods typically falter \cite{Yun2012,Ji2014}.  Skeleton-based   action recognition   has evolved  from early handcrafted features and manifold-based representations \cite{Yun2012, refref18, Ji2014, Li2015a, refref40, refref41, refref59, refref39, Vemulapalli2014, refref11, RiemannianManifoldTraject, refref61, Huangcc2017, ref23} to deep temporal architectures such as LSTMs and GRUs \cite{Du2015, Liu2016, Zhua2016, Zhang2017b, GCALSTM, DeepGRU}.  Subsequently,   attention-based models, including Graph Convolutional Networks (GCNs),   have set new benchmarks by directly exploiting the natural kinematic topology of the human skeleton to capture intricate spatial and temporal dependencies \cite{Jiang2020, Song2017, Liu2021}.\\

\indent Despite these architectural advancements, the performance of state-of-the-art models remains bottlenecked by a heavy reliance on massive, expert-labeled datasets. The manual annotation of skeletal sequences is notoriously labor-intensive,   prompting the exploration of mitigative strategies such as data augmentation \cite{DA2}, few-shot learning \cite{TL1}, and self-supervised pre-training \cite{SS1}. While useful, these methods often rely on proxy tasks that cannot fully substitute for the discriminative power of ground-truth labels. Active Learning (AL) \cite{Burr2009} offers a more direct trajectory toward data efficiency by tasking an ``oracle'' with labeling {\it only} the most informative samples, typically prioritized via uncertainty \cite{SS1, Culotta2005, zhao2023uncertainty, kim2023batch} or diversity-based criteria \cite{Ref50a, Ref51a, wu2022adaptive, zhang2022multi}. However, a critical limitation persists: most contemporary AL approaches are confined to sampling from a fixed pool, lacking a rigorous framework for data ``design.'' This reliance on static pools often yields suboptimal datasets that fail to span the underlying data manifold \cite{du2022efficient, cai2022self, kondo2023active}. \\

\indent  In this paper, we propose a paradigm shift from pool-based selection to {\it exemplar design}. Our first contribution is a  probabilistic AL model that builds  candidate data for labeling as the optimal solution to an  objective function balancing  manifold exploration with classifier uncertainty. However, deriving these exemplars directly within the highly nonlinear skeletal space is inherently intractable.  Without a mechanism to explicitly navigate the complex data manifold,  optimization offers no guarantee that resulting exemplars will remain in-distribution.  Our second contribution addresses this tractability gap through the integration of stable and invertible  GCNs.   To establish a stable, bijective mapping between  skeletal data and the  latent space, we introduce the \textit{R-Cayley reparameterization}—a novel relaxation of the standard Cayley transform that moves beyond the rigid isometry of  orthonormal weights. While strict orthonormality ensures stability, it restricts the network's  expressive power  across skeletal manifolds. By regulating the condition number of the trained GCNs, our approach enforces bi-Lipschitzian continuity  while allowing the GCNs  to capture a broader, more expressive class of transformations. This ensures that the  exemplars are designed within tractable latent distribution and remain strictly grounded on the data manifold upon back-projection in the ambient space,  ultimately yielding accurate, label-frugal GCNs for skeleton-based  recognition.

\section{Our  Exemplar Design}

Our AL framework integrates two interacting components: a \textit{display model} for targeted data acquisition and a \textit{learning model} for label-frugal classification.  Given an unlabeled dataset  $\mathcal{U} = \{\mathbf{x}_1, \dots, \mathbf{x}_N\} \subset \mathbb{R}^d$, each iteration $t \in \{0, \dots, T-1\}$ constructs a subset $\mathcal{D}_t$ --- the \textit{display} --- comprising designed  exemplars derived from $\mathcal{U}$. Upon receiving labels $\mathcal{Y}_t \subseteq\{1,\dots,K\}$ from an oracle, the classifier $f_t$ is updated on the cumulative dataset $\cup_{k=0}^t(\mathcal{D}_k, \mathcal{Y}_k)$. Our primary contribution \textit{flexibly designs displays} rather than merely sampling static pools of data.  By designing  exemplars that maximally challenge the current classifier $f_t$, we elicit high-utility data and labels, facilitating a more robust estimation of $f_{t+1}$ in subsequent AL iterations. \\

We consider a probabilistic framework that assigns a \textit{membership degree} $\boldsymbol{\mu}_i$ to each unlabeled data, quantifying its   contribution to the designed  display $\mathcal{D}_{t+1}$. Defining $\boldsymbol{\mu} \in \mathbb{R}^N$ as the vector of these memberships $\{\boldsymbol{\mu}_i\}_i$,  an  exemplar $\mathbf{x}^* \in \mathcal{D}_{t+1}$ is formulated as a convex combination of unlabeled samples $\mathbf{X}_{\mathcal{U}} \in \mathbb{R}^{N \times d}$ as   
$\mathbf{x}^*(\boldsymbol{\mu}) = \mathbf{X}_{\mathcal{U}}^{\top} \boldsymbol{\mu}$ (with $\sum_{i}\boldsymbol{\mu}_i=1$).  We find the optimal $\boldsymbol{\mu}$ by solving the following constrained minimization problem:
\begin{equation}\label{loss} 
\begin{array}{c}
\displaystyle \min_{\scriptsize \begin{array}{c} \boldsymbol{\mu}\ge\mathbf{0}, \\ \|\boldsymbol{\mu}\|_1 = 1\end{array}}
\underbrace{\left(\mathbf{S}^{\top} \boldsymbol{\mu}\right)^{\top} \log \left(\mathbf{S}^{\top} \boldsymbol{\mu}\right)}_{\text{Uncertainty}} + \alpha \underbrace{\mathbf{1}_{\mathcal{L}}^{\top} \mathbf{K}_\sigma(\boldsymbol{\mu})}_{\text{Diversity}} \\  + \beta \underbrace{\boldsymbol{\mu}^{\top} \log \boldsymbol{\mu}}_{\text{Entropy}}. 
\end{array}
\end{equation}
In this objective: (i) $\mathbf{S}$ is the $N \times K$ softmax (data vs.  classes) prediction matrix, (ii) $\mathbf{1}_{\mathcal{L}}$ is a vector of ones for the cumulative labeled data $\mathbf{X}_{\mathcal{L}}$, and (iii) $\mathbf{K}_\sigma(\boldsymbol{\mu}) \in \mathbb{R}^{|\mathbf{X}_{\mathcal{L}}|}$ is a vector whose entries $\{[\mathbf{K}_\sigma(\boldsymbol{\mu})]_\ell = \exp\big(-\|\mathbf{x}^*(\boldsymbol{\mu}) - \mathbf{x}_l \|_2^2 \slash (2\sigma^2)\big)\}_\ell$ measure the similarity between the new designed exemplar and previously labeled ones.
The \textit{uncertainty} term reaches its minimum when $\mathbf{x}^*(\boldsymbol{\mu})$ is positioned in regions of maximal classifier ambiguity. The \textit{diversity} term drives exploration by prioritizing exemplars  far from the previously labeled ones. Finally, the \textit{entropy} term regularizes the solution,  and also  enables  a closed-form  fixed-point solution. The gradient optimality conditions yield the following iterative solution:
\begin{equation}\label{eq2}
\begin{array}{l}
\displaystyle \boldsymbol{\mu}^{(\tau +1)} := \frac{\hat{\boldsymbol{\mu}}^{(\tau +1)}}{\|\hat{\boldsymbol{\mu}}^{(\tau +1)} \|_1}, \\  \displaystyle \hat{\boldsymbol{\mu}}^{(\tau +1)} = \exp \left\{ -\frac{1}{\beta} \left[ \mathbf{G}_{\text{Unc}}^{(\tau)} + \alpha \mathbf{G}_{\text{Div}}^{(\tau)} \right] \right\},
\end{array}
  \end{equation}
where 
  \begin{equation}\label{eq4} 
\mathbf{G}_{\text{Unc}} = \mathbf{S} \left( \log \left( \mathbf{S}^{\top} \boldsymbol{\mu} \right) + \mathbf{1}_K \right),
\end{equation}
and 
\begin{equation}\label{eq5} 
\mathbf{G}_{\text{Div}} = \frac{1}{\sigma^2} \mathbf{X}_{\mathcal{U}} \left[ \mathbf{X}_{\mathcal{U}}^{\top}( \boldsymbol{\mu} \mathbf{1}_{\mathcal{L}}^{\top} \mathbf{K}_\sigma(\boldsymbol{\mu})) - \mathbf{X}_{\mathcal{L}}^{\top} \mathbf{K}_\sigma(\boldsymbol{\mu}) \right].
\end{equation}
In this solution, $\mathbf{G}_{\text{Unc}}$ controls data  weights $\boldsymbol{\mu}$ in $\mathbf{x}^*(\boldsymbol{\mu})$  based on  classifier uncertainty; the component $\mathbf{S} \log \mathbf{S}^\top \boldsymbol{\mu}$ acts as a corrective force that decays the membership of data  in over-represented or low-entropy regions to focus on ambiguous boundaries. Simultaneously, $\mathbf{G}_{\text{Div}}$ functions as a \textit{repulsive potential field}, pushing the membership mass (in $\boldsymbol{\mu}$) away from ``covered'' regions toward unexplored data manifolds. \\

\noindent In order to  ensure the model is self-adaptive, $\sigma$ in $\mathbf{K}_\sigma(\boldsymbol{\mu})$ is dynamically adjusted by taking the expectation (with respect to labeled data) of the terms inside  $\mathbf{K}_\sigma$, i.e., $2\sigma^2 = \frac{\alpha}{|\mathbf{X}_{\mathcal{L}}|} \mathbf{1}_{\mathcal{L}}^\top \log(\mathbf{K}_{1}(\boldsymbol{\mu})^{-1})$; setting $\sigma$ proportionally to $\alpha$ allows the latter to be absorbed in Eq. \ref{eq2}, thereby minimizing the  number of hyperparameters. In Eq. \ref{eq2}, $\beta$ serves as an iteration-dependent scaling factor that controls the exponential function in $\boldsymbol{\mu}$, defined as the reciprocal of the input norm: $\beta = 1/\|\log(\hat{\boldsymbol{\mu}}^{(\tau+1)})\|_1$. Finally, the normalization step (Eq. \ref{eq2}) enforces $\|\boldsymbol{\mu}\|_1 = 1$ so as to maintain a probability distribution. Starting from random $\boldsymbol{\mu}^{(0)}$, the fixed-point process typically converges to an optimal solution $\tilde{\boldsymbol{\mu}}$ within few iterations.

\section{Our Cayley-Reparameterized GCNs} \label{model}
\def\SS{{\cal S}}
\def\E{{\cal E}}
\def\A{{\bf A}}
\def\F{{\cal F}}
\def\UU{{\bf U}} 
\def\W{{\bf W}}
\def\WW{{\bf W}}
\def\N{{\cal N}}

The effectiveness of our AL framework hinges on the fidelity of the designed exemplars; the latter must align with the underlying data distribution. However, direct optimization in the ambient space is problematic when data resides on arbitrary nonlinear manifolds. Without explicit manifold parameterization, the optimization in Eq. \ref{loss} cannot ensure that generated exemplars remain on-manifold, often leading to out-of-distribution data and degraded classification. To resolve this, we introduce our Cayley-Reparameterized  GCN architecture to shift exemplar generation to a  latent domain; as shown subsequently and later in experiments, optimizing within this  latent space provides a significant performance boost over direct ambient-space exemplar design.

\subsection{Stable \& Invertible GCNs} 
Formally, for a graph $\mathcal{G}=(\mathcal{V}, \mathcal{E})$ with signal $\mathbf{U} \in \mathbb{R}^{s \times n}$ and adjacency matrix $\mathbf{A} \in \mathbb{R}^{n \times n}$, GCNs learn a bank of $C$ filters $\mathbf{W} \in \mathbb{R}^{s \times C}$ to define the convolution $(\mathcal{G} \star \mathcal{F})_{\mathcal{V}} = g(\mathbf{A} \mathbf{U}^\top \mathbf{W})$, where $g(\cdot)$ is a nonlinear activation. Here, $\mathbf{A}$ acts as an operator that aggregates local neighborhood representations, while $\mathbf{W}$ projects these aggregated representations into a $C$-dimensional feature space. In what follows, these multiple two-stage processes are formally subsumed as a $L$-layered network $f$ parameterized by $\theta = \{ \mathbf{W}_1, \dots, \mathbf{W}_L \}$, where $\mathbf{W}_\ell \in \mathbb{R}^{d_{\ell-1} \times d_\ell}$ represents the weight tensor for the $\ell^{\text{th}}$ layer. For $\ell \in \{2, \dots, L\}$, the layer output is $\phi^\ell = g_\ell({\mathbf{W}_\ell}^\top \phi^{\ell-1})$, where $g_\ell$ is the activation function and (without loss of generality) the bias term is omitted.  Our framework requires GCNs to be both invertible (bijective) and stable.   The invertibility of $f$ allows us to explicitly recover  ambient data via the recursive relation $\phi^{\ell-1} = (\mathbf{W}_\ell^\top)^{-1} g_\ell^{-1}(\phi^\ell)$ whilst  stability further ensures that the inverse mapping $f^{-1}$ is well-behaved; specifically, when $f^{-1}$ is evaluated on an output (latent) distribution, the resulting input (ambient) reconstructions remain anchored to the original data manifold. This can be guaranteed by constraining $f$ to be $KM$-Lipschitzian, i.e.,    $f$ and $f^{-1}$ are respectively  $K$- and $M$-Lipschitz continuous. \\

While minimizing the product  $KM$ is  challenging \cite{heinonen2005lectures},  our GCN design,  described subsequently,  allows both constants to be tightly bounded.  More precisely,  if (i) the activation functions $\{g_\ell(\cdot)\}_{\ell=2}^L$ are bijective; (ii) their derivatives satisfy $l \leq |g'_\ell(\cdot)| \leq u$; and (iii) the condition numbers of the weight matrices in $\theta$ are  bounded by $\kappa$, then one may show that  the bidirectional network $f$ is $KM$-Lipschitz continuous with $KM =  \left( {\kappa u}/{l} \right)^{L-1}$ \footnote{In practice, $u=0.99$,  $l=0.95$ are  Leaky-ReLU slopes,  set  to make $KM$ small.}.  The feasible parameter space that makes $KM=1$---and thereby our GCNs the most stable (1-Lipschitzian)---corresponds to the \textit{Stiefel Manifold}, denoted as $\mathcal{V}_{d_\ell}(\mathbb{R}^{d_{\ell-1}}) = \{ \mathbf{W}_\ell \in \mathbb{R}^{d_{\ell-1} \times d_\ell} : \mathbf{W}_\ell ^\top \mathbf{W}_\ell  = \mathbf{I} \}$; hence, $\{\mathbf{W}_\ell\}_\ell$  are found  as  
$\textrm{argmin}_{\{\mathbf{W}_\ell \in \mathcal{V}_{d_\ell}(\mathbb{R}^{d_{\ell-1}})\}_\ell} \text{CE}(f; \{\mathbf{W}_\ell\}_\ell)$,
\noindent where $\mathbf{I}$ is the identity matrix, and $\text{CE}$ stands for cross-entropy.   While  strict orthonormality ensures $\kappa=1$ and minimizes the $KM$ constants, it introduces significant architectural rigidity. Such ``hard'' constraints can stifle the network's expressivity, as the model is restricted to capturing only near-isometric transformations between the ambient and latent spaces  (due to  Leaky-ReLU).  To enhance the generative and discriminative capacity of the network, we must allow for a broader class of transformations as shown subsequently.

\subsection{Our Cayley-Reparameterized GCNs}\label{optim}
To establish a stable, invertible mapping between ambient and latent spaces without sacrificing expressivity, we introduce the \textit{Relaxed Cayley (R-Cayley) reparameterization}. While standard Cayley transform \cite{cayley2021} enforces strict isometry via orthonormal weights, they constrain the network's capacity to model near-isometric manifold transformations only.  We  address this limitation by reparameterizing the weights $\mathbf{W}_\ell$ using a single unconstrained latent matrix $\mathbf{B}_\ell \in \mathbb{R}^{d_{\ell-1} \times d_{\ell}}$ as:
\begin{equation}\label{eq:rcayley}
\begin{array}{l}
  \mathbf{W}_\ell = \underbrace{\left(\exp(\text{diag}(\mathbf{B}_\ell)) + \lambda \mathbf{I}\right)}_{\text{Anisotropic Scaling }} \\  \quad \quad \quad \quad \quad \cdot \underbrace{(\mathbf{I} + \text{skew}(\mathbf{B}_\ell))(\mathbf{I} - \text{skew}(\mathbf{B}_\ell))^{-1}}_{\text{Orthonormal Transformation }},
  \end{array}
\end{equation}
where $\text{skew}(\mathbf{B}_\ell) = \frac{1}{2}(\mathbf{B}_\ell - \mathbf{B}_\ell^\top)$ corresponds to  the skew-symmetric transform of  $\mathbf{B}_\ell$ and $\lambda>0$.  This   formulation unifies scaling and rotation; diagonal entries $\exp(\text{diag}(\mathbf{B}_\ell))$ control anisotropic scaling whereas the remaining off-diagonal entries dictate the orthonormal transformation via $\text{skew}(\mathbf{B}_\ell)$, whose term $\mathbf{I} - \text{skew}(\mathbf{B}_\ell)$ is unconditionally invertible for real skew-symmetric matrices. By disentangling rotation and scaling within a single matrix $\mathbf{B}_\ell$, our R-Cayley formulation   enhances  expressivity while guaranteeing  bi-Lipschitz stability  as shown subsequently and through experiments. \\

\begin{proposition}\it
Given a bidirectional network $f$ of depth $L$, let the following conditions hold: (i) The pointwise activation functions $\{g_\ell(\cdot)\}_{\ell=2}^L$ are bijective with derivatives $l \leq |g'_\ell(\cdot)| \leq u$. (ii) Each weight matrix $\mathbf{W}_\ell$ is reparameterized via Eq. \ref{eq:rcayley}.  Then,  the network $f$ is $KM$-Lipschitz continuous with
$KM = \prod_{\ell=2}^L \left( \frac{s_{\ell, \max} + \lambda}{s_{\ell, \min} + \lambda} \cdot \frac{u}{l} \right)$,
where  $s_{\ell, \max}$ and $s_{\ell, \min}$ respectively denote the maximum and minimum values of the diagonal entries of $\exp(\text{diag}(\mathbf{B}_\ell))$. 
\end{proposition}

\noindent In view of space,  details of the proof are omitted.   More importantly,   the stability of the bidirectional mapping is  determined by the \textit{spectral spread} of $\mathbf{B}_\ell$ moderated by $\lambda$. This highlights a fundamental trade-off: a large $\lambda$  dampens the  influence of the learned anisotropy, pushing the network toward a stable but near-isometric (less expressive) regime.  A small $\lambda$ ---together with the variances in $\text{diag}(\mathbf{B})$---  provides the anisotropic (expressive) flexibility required to capture highly nonlinear mapping between latent and ambient spaces. With the appropriate $\lambda$-setting  (see later experiments), the R-Cayley reparameterization inherently  ensures a {\it stable bijective mapping} that allows for the precise search of exemplars in the latent space and their subsequent back-projection to the ambient space while remaining on-manifold.

\section{Experiments}\label{sec:experiments}

To evaluate the performance of our framework, we conduct action recognition experiments on two standard skeleton-based benchmarks: SBU  \cite{Yun2012} and FPHA \cite{refref11}. SBU consists of 282 sequences of dyadic actions, where two actors perform one of eight interaction categories. Each frame captures the 3D coordinates of 15 joints per person via a Microsoft Kinect sensor. We follow the experimental protocol and train-test split established in \cite{Yun2012}. For the FPHA dataset, the task involves 1,175 sequences across 45 hand-action classes, representing significant variability in execution speed, scale, and camera viewpoint. Each skeleton sequence tracks the temporal trajectory of 21 hand joints in 3D space. Adhering to the evaluation pipeline in \cite{refref11}, we employ a 1:1 split, using 600 sequences for training and 575 for testing. \\ 
\begin{table}  
\centering
\resizebox{0.7\columnwidth}{!}
{

 \begin{rotatebox}{90}{
\begin{tabular}{c|cc|c}
{\bf \rotatebox{-0}{Data}}      &  {\bf  \rotatebox{-0}{Method}}      &   & {\bf  \rotatebox{-0}{Acc (\%)}}\\
\hline 
\hline
    \multirow{13}{*}{\rotatebox{-90}{SBU}}    & Raw Position \cite{Yun2012} & $ \ $   & 49.7   \\ 
  & Joint feature \cite{Ji2014}  & $ \ $   & 86.9   \\
  & CHARM \cite{Li2015a}       & $ \ $    & 86.9   \\

&H-RNN \cite{Du2015}         & $ \ $    & 80.4   \\ 
&ST-LSTM \cite{Liu2016}      & $ \ $    & 88.6    \\ 
&Co-occurrence-LSTM \cite{Zhua2016} & $ \ $  & 90.4  \\ 
&STA-LSTM  \cite{Song2017}     & $ \ $   & 91.5  \\ 
&ST-LSTM + Trust Gate \cite{Liu2016} & $ \ $  & 93.3 \\
&VA-LSTM \cite{Zhang2017}      & $ \ $  & 97.6  \\
 &GCA-LSTM \cite{GCALSTM}                    &   $ \ $      &  94.9     \\ 

&Riemannian manifold. traj~\cite{RiemannianManifoldTraject} &  $ \ $  & 93.7 \\
&DeepGRU  \cite{DeepGRU}        &    $ \ $   &    95.7    \\
&RHCN + ACSC + STUFE \cite{Jiang2020} & $ \ $   & 98.7 \\ 

  & \textcolor{black}{Our baseline GCN} &              &        98.4      \\
  \hline \hline
  \multirow{21}{*}{\rotatebox{-90}{FPHA}} &    2-stream-color \cite{refref10}    & $ \ $   &  61.56 \\
& 2-stream-flow \cite{refref10}   & $ \ $   &  69.91 \\  
 &2-stream-all \cite{refref10}      & $ \ $  &  75.30 \\

&HOG2-dep \cite{refref39}        & $ \ $ &  59.83 \\    
&HOG2-dep+pose \cite{refref39}  & $ \ $    &  66.78 \\ 
&HON4D \cite{refref40}             & $ \ $   &  70.61 \\ 
&Novel View \cite{refref41}       & $ \ $    &  69.21  \\ 

&1-layer LSTM \cite{Zhua2016}    & $ \ $      &  78.73 \\
&2-layer LSTM \cite{Zhua2016}    & $ \ $    &  80.14 \\ 

&Moving Pose \cite{refref59}      & $ \ $    &  56.34 \\ 
&Lie Group \cite{Vemulapalli2014}  & $ \ $  &  82.69 \\ 
&HBRNN \cite{Du2015}             & $ \ $     &  77.40 \\ 
&Gram Matrix \cite{refref61}      & $ \ $     &  85.39 \\ 
&TF    \cite{refref11}             & $ \ $     &  80.69 \\  

&JOULE-color \cite{refref18}     & $ \ $    &  66.78 \\ 
&JOULE-depth \cite{refref18}     & $ \ $     &  60.17 \\ 
&JOULE-pose \cite{refref18}      & $ \ $   &  74.60 \\ 
&JOULE-all \cite{refref18}        & $ \ $    &  78.78 \\ 

&Huang et al. \cite{Huangcc2017}  & $ \ $     &  84.35 \\ 
&Huang et al. \cite{ref23}        & $ \ $     &  77.57 \\  

&HAN  \cite{Liu2021}  & $ \ $ & 85.74 \\

&Our baseline GCN         & $ \ $       & 88.17 \\ 
\hline

\end{tabular}
 } \end{rotatebox}
 }
 \caption{Comparison of our baseline GCN against related work on the SBU and  FPHA databases (better to zoom the PDF).}\label{compare2}
\end{table}

\noindent{\bf Implementation details and baseline architectures.} Our GCN models are optimized over 2700 epochs using  Adam  with a momentum of 0.9. We set the batch size to 200 for  SBU  and 600 for FPHA, and we implement an adaptive learning rate $\nu$, governed by loss derivative.   Specifically, $\nu$ is decayed by a factor of 0.99 when the derivative increases and scaled by $0.99^{-1}$ otherwise. All experiments have been conducted on a GeForce GTX 1070 GPU (8 GB), and neither dropout nor data augmentation have been used. For   SBU, the architecture consists of three sequential blocks, each integrating a single-head attention mechanism and a convolutional layer with 8 filters. For  FPHA, we use a similar architecture, but we scale the capacity of the model by increasing the number of filters from 8 to 16. Table~\ref{compare2} shows baseline GCNs trained in  a fully supervised setting. Our objective is to narrow the gap,  as much as possible,  relative to  these upper bound performances whilst being label-frugal.\\

\begin{table}[h]
 \begin{center}
\resizebox{0.65\columnwidth}{!}{
  \begin{tabular}{clclclc}   
    \rotatebox{0}{Labeling rates} & \multicolumn{2}{c}{\rotatebox{0}{{Accuracy}}} & \rotatebox{0}{Observation} \\ \hline
                     & \rotatebox{0}{SBU} & FPHA &  \\ 
 \hline
  \hline
    100\%    &    98.40    &   88.17   & Baseline GCN (not label-efficient)\\
     \hline
    \multirow{5}{*}{\rotatebox{0}{15\%}}      & 60.00  &  40.52 & wo display model (random display)  \\
                                  & \underline{64.61}  & \underline{47.21}  & + display model + ambient  (our) \\
                                  &   \bf80.00   & \bf 51.21  & + display model + latent  (our) \\
                                  &46.15  & 41.73  & uncertainty (margin-based) \\                                                                                                                                          & 60.00  &  46.26  & diversity (coreset-based)\\                                        
    \hline
      \multirow{5}{*}{\rotatebox{0}{7\%}}   & \underline{56.92}  & 32.21         & wo display model (random display)  \\
                                             & 41.53  &  \underline{37.53}          &  + display model + ambient  (our) \\
                                  & \bf64.61  &  \bf39.12           &  + display model + latent  (our) \\                 
                                             & 30.76  &     33.17   & uncertainty (margin-based) \\                                                                                                                                                                              & 55.38   & 36.77  &  diversity (coreset-based)\\                                        
    \hline 
        \multirow{5}{*}{\rotatebox{0}{3\%}} & 29.23 &  21.11   & wo display model (random display) \\
                                             & \underline{32.30}  &     \underline{24.60}    & + display model + ambient   (our) \\
                                  & \bf35.38  &   \bf25.64 & + display model + latent  (our) \\
                                             & 29.23  &  21.74     & uncertainty (margin-based) \\                                                                                                                                                                              & 29.23   &   24.10 & diversity (coreset-based) \\                                        
      \hline 
 
  \end{tabular}}
\end{center}
\caption{This table shows detailed performances and ablation study on SBU and FPHA for different  labeling rates. Here ``wo'' stands for ``without'',   and ``latent (our)'' refers to our  R-Cayley reparam with $\lambda=1$.  Best results are shown in bold and second best results underlined.}\label{table21}
\end{table}

\noindent {\bf Display model:  analysis and ablation study.} Table~\ref{table21} shows the performance of our framework on  SBU and FPHA respectively. Direct application of our exemplar model within the ambient space yields high classification accuracy, frequently surpassing existing selection strategies. However, the most significant performance gains are observed when leveraging the latent space, demonstrating a powerful synergy between our display model and the learned latent representations. When benchmarked against alternative display selection strategies---including random, diversity-based  \cite{zhang2022multi}, and uncertainty-based selection \cite{zhao2023uncertainty}---our method consistently achieves superior results across most of the labeling regimes. The advantages of our approach are most pronounced in these data-scarce scenarios; while random  remains competitive at relatively high  labeling rates  --- consistent with previous findings \cite{Burr2009} --- its effectiveness diminishes considerably at extreme low rates (3\%). In these regimes, random  fails to capture the underlying data distribution effectively, necessitating the  manifold-aware selection  provided by our framework. While uncertainty  prioritizes low-confidence data,  they often lack  richness; conversely, diversity  fails to refine  classifier's  boundaries.  Unlike these static approaches that are restricted to selecting from a pre-defined pool, our display model learns adaptive exemplars within  latent spaces of stable, invertible GCNs. This allows more expressivity, resulting in better performance even under extreme label frugality.\\

\noindent {\bf R-Cayley  versus   other regularizations.}   Table~\ref{abla}  showcases the impact of {R-Cayley reparameterization} across various $\lambda$-settings, benchmarking it against a baseline of {no regularization} (config. \#1), {Condition Number (CN) regularization} (config. \#2,  which penalizes the  product of spectral norms of the weight matrices and their inverses in order  to stabilize transformations), and strict {Orthogonality (OR) regularization} (config. \#3,  via standard Cayley).  The results consistently demonstrate the superiority of R-Cayley when $\lambda$ is optimally tuned (config. \#6 vs. \#4, \#5, \#7), revealing a decisive advantage over strict orthonormality (config. \#6 vs. \#3). By  trading rigid stability for network flexibility, R-Cayley still maintains a controlled condition number and competitive Fr\'echet Inception Distance (FID) while simultaneously enhancing classification accuracy. We observe that while increasing $\lambda$ bolsters network stability and exemplar quality (config. \#6), overestimating this parameter (config. \#7) induces extra  rigidity that hinders cross-entropy minimization. Conversely, an underestimated $\lambda$ (configs. \#4, \#5) provides excessive model flexibility,  resulting in higher FID and CN scores that suggest the generation of out-of-distribution exemplars which  impact   GCN generalization.  Ultimately, an intermediate $\lambda$ (config. \#6) strikes the optimal balance, maximizing the effectiveness of the reparameterization by reconciling discriminative power with {manifold  preservation}.

\begin{table}\centering
 \resizebox{0.75\columnwidth}{!}{
 \begin{tabular}{c|c||cccc}
Dataset & Regularizer              & Acc $\uparrow$ & Observed CN  $\downarrow$& FID Score $\downarrow$ & config\\
   \hline
  \multirow{8}{*}{\rotatebox{0}{SBU}}    & None                     &  67.69   &  $2.45 \times 10^3$   &   $1.84 \times  10^{18}$  & \#1 \\
&Condition Number      &  60.00  & 26.83   & 234.72    & \#2 \\
&Strict Ortho (Cayley)        &  52.30   &  1.00  &  1.10   & \#3  \\
&R-Cayley  ($\lambda=0.01$)        &  53.84  &  7.07 &  3.49  & \#4    \\   
&R-Cayley  ($\lambda=0.10$)        &  41.53  & 7.41  &  \underline{1.48} & \#5 \\  
&R-Cayley  ($\lambda=1.00$)        &  \bf80.00  & \bf3.89  & \bf1.11  & \#6    \\   
&R-Cayley  ($\lambda=2.00$)        &  \underline{67.69}  & \underline{4.14} &  2.77  & \#7    \\   
 \hline
 \hline
  \multirow{8}{*}{\rotatebox{0}{FPHA}}    &  None                     & 16.81   &  $3.29 \times 10^5$   &  $1.34 \times  10^{15}$  & \#1 \\
&Condition Number      &  42.48  & 8.92   & 2.41    & \#2 \\
&Strict Ortho (Cayley)        &  16.81  &  1.00  &  2.15  & \#3  \\
&R-Cayley  ($\lambda=0.01$)        &  26.08  &  13.93   &  3.17  & \#4    \\   
&R-Cayley  ($\lambda=0.10$)        &  21.02  &   12.75    & 2.84 & \#5 \\  
&R-Cayley  ($\lambda=1.00$)        & \bf51.21  &  \underline{10.09} & \bf1.63  & \#6    \\   
&R-Cayley  ($\lambda=2.00$)        &  \underline{42.48}    &   \bf9.99 &  \underline{1.81}  & \#7    \\   
 \hline
\end{tabular}}
\caption{\textcolor{black}{This table shows the impact of  R-Cayley reparameterization  (for different setting of $\lambda$)  and its comparison against different regularizations,  used to train the GCNs,   including the challenging condition number (CN) minimization,  and strict orthonormality (OR); OR is obtained from the original Cayley as discussed in section~\ref{optim}.  Comparison is also provided when no regularization is used (dubbed as None); this corresponds to a GCN trained without any constraints on its weights.   All these results correspond to the highest labeling regime of  our experiments  (i.e.,  15\%). }}\label{abla} 
 \end{table} 

\section{Conclusion} 

This paper introduces a label-frugal framework for skeleton-based action recognition that balances model expressivity with bi-Lipschitz stability. Using a Relaxed-Cayley reparameterization within an invertible GCN architecture, we capture the intricate geometry of skeletal motion while enforcing stability which renders the design of optimal training exemplars  more tractable.  Our  display  selection model leverages this stable GCN design to identify informative exemplars by minimizing a unified objective of diversity and uncertainty. Extensive experiments on the SBU and FPHA datasets show that our framework consistently outperforms related selection strategies, particularly in label-limited regimes.


\begin{thebibliography}{1}
\bibitem{aaaaa9}  H. Sahbi.  A particular Gaussian mixture model for clustering and its application to image retrieval.   Soft Computing 12 (7), 667-676
\end{thebibliography}
\end{document}